# A Learning oriented DLP System based on Classification Model

KISHU GUPTA[1]
ASHWANI KUSH[2]

[1]Department of Computer Science & Applications,
Kurukshetra University,
India
[2]Institute of Integrated and Hons. Studies,
Kurukshetra University,
India
[1]kishugupta2@gmail.com
[2]akush20@gmail.com

**Abstract:** Data is the key asset for organizations and data sharing is lifeline for organization growth; which may lead to data loss. Data leakage is the most critical issue being faced by organizations. In order to mitigate the data leakage issues data leakage prevention systems (DLPSs) are deployed at various levels by the organizations. DLPSs are capable to protect all kind of data i.e. DAR, DIM/DIT, DIU. Statistical analysis, regular expression, data fingerprinting are common approaches exercised in DLP system. Out of these techniques; statistical analysis approach is most appropriate for proposed DLP model of data security.

This paper defines a statistical DLP model for document classification. Model uses various statistical approaches like TF-IDF (Term Frequency- Inverse Document Frequency) a renowned term count/weighing function, Vectorization, Gradient boosting document classification etc. to classify the documents before allowing any access to it. Machine learning is used to test and train the model. Proposed model also introduces an extremely efficient and more accurate approach; IGBCA (Improvised Gradient Boosting Classification Algorithm); for document classification, to prevent them from possible data leakage. Results depicts that proposed model can classify documents with high accuracy and on basis of which data can be prevented from being loss.

**Keywords:** BoW; False positive/negative rate; IGBCA; Sparse matrix; TF-IDF; Vectorization

## 1. Introduction

The NIST explains computer security as "protection afforded to an automated information system in order to attain the applicable objectives of preserving the integrity, availability, and confidentiality of information system resources (includes hardware, software, firmware, information/data, and telecommunications)" [1].

Data leakage is access to organizational/individual sensitive data by some unauthorized party. Data leakage may result in severe problems for organization or individual. Inadequate availability of computational resources leads to data sharing among diverse entities to fulfil the organization goals [2], [3]. Enterprises share data with other entities for rapid growth, more productivity and profit. Various nation shares defence sector data for cooperation and coordination in war time, to learn war tactics of their allies to combat enemies easily and effectively in high time. Medical professional also shares sensitive data to avail latest and effective treatment. Leakage can compromise the enterprise plans, patient records, financial details; defence secrets which affect healthy competition among organizations, confidentiality of patient and defence security details and also accounts security. A report 'Data Loss Statistics' specifies that the data leaks are incremental in nature in respect of data leakage size and its impact [4]. For example, well-known online shopping platform eBay faced one of the largest data leak, in which critical data like name, email address and some other vital personal information of about 145 million users was leaked which hampered its functionality ( a direct loss) and also many customers migrated to other platform due to loss of trust (an indirect loss) [5]. One more instance of data leakage is Sony PlayStation account details breaching of about 77 million users, followed by entire shut-down of PlayStation network for a very long duration and also a detailed public apology by CEO of company.

Data leakage may happen from within or even outside the organization premises. Data Leakage Prevention systems (DLPSs) are recommended as the most preferable way to resolve the crisis of data leakage. DLP is implemented with goal to avoid data leak. Popular standard approaches like Firewalls, IDS (Intrusion Detection System), Anti malware etc. are based on DPI i.e. deep packet inspection whereas DLP performs DCA i.e. deep content analysis [6]. DCA continuously monitors data and find out the content which seems sensitive for an organization. DLP can efficiently handle all types of data i.e. Data 'in transit/motion' (travelling on network internally or externally), 'at rest' (data stored in repository) and 'in use' (in access by users) data [6]. As per [7] DLP is:"*a system that is designed to detect and prevent the unauthorized access, use, or transmission of confidential information*". DLP described according to [8] is:"*Products that, based on central policies, identify, monitor, and protect data at rest, in motion, and in use, through deep content analysis*". DLP can allow, prevent, restrict or completely block the data access depending on data states [9], [10].

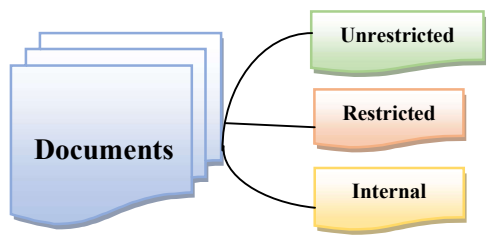

**Figure 1:** Document classes

Proposed DLP solution classifies data as unrestricted, restricted or internal classes as shown in Figure 1 to provide access to data accordingly.

DLP analyses both the content as well as context of documents. Context analysis covers scrutiny of various attributes like size, time, format and sender/receiver of data whereas regular expressions, fingerprinting, statistical approaches etc. are used to carry out content analysis [11]. On basis of both of these analysis DLPSs take any action like alert, allow, block, encrypt [12] etc. Figure 2 explains the DLP solution layout.

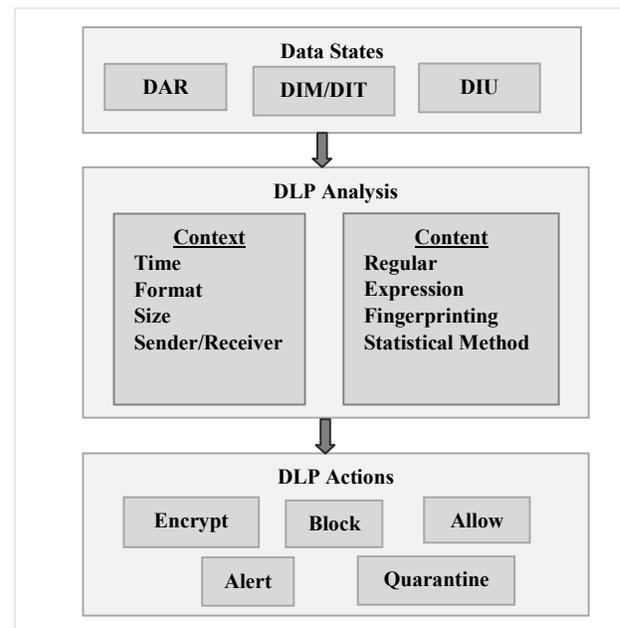

**Figure 2:** DLP Solution Layout [6]

## 1.1. Machine Learning Framework

Data collected and stored in repositories is available in text form. Machine learning models cannot process raw text directly, instead ML can deal only with numerical values; specifically, vectors of numbers [13]. In order to reflect various linguistic properties of the text, vectors of number are derived from textual data by performing text preprocessing using language processing like NLP [14], [15]. This is called feature extraction or feature encoding [16]. Further to design the vocabulary of words, a well recognized and easy technique for feature extraction from textual data named as bag-of-words is used. BoW approach is very simple and flexible method used to extract features from the content. A BoW is like representative of text data which define the occurrence of various words in the document. Each known word occurring in vocabulary is known as token/lamma. To create vectors of number from textual data vectorization is carried out [17]. Every unique word in the textual data is corresponded with a number. Thus all the raw textual data i.e. sequence of characters transformed in vectors i.e. sequences of numbers. After obtaining vectors of number word scoring [18], [19], [20] is carried out. TF-IDF is one of best method for this purpose. It finds out the importance of a word for a document as more frequently occurring words may get high score and can be considered as sensitive despite they are not sensitive actually. A ML approach usually consists of two phases: training and testing. To train model and then to check model performance data is portioned in two sets namely the train set and the test set. Proposed model uses GBCA/IGBCA for training/tuning purpose. Figure 3 explains the machine learning framework.

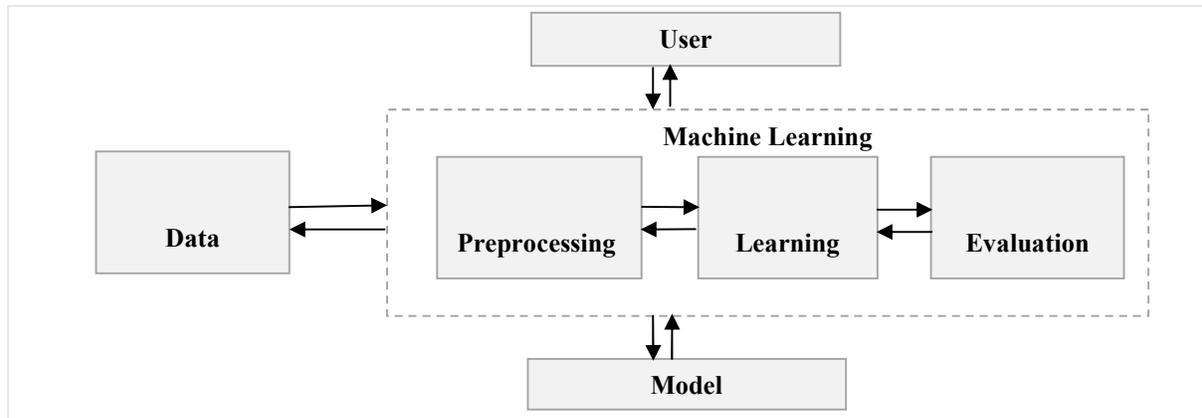

**Figure 3:** Framework of Machine Learning

Machine learning generally encompass following steps:
- Extract features i.e. attribute and data classes from training set
- Perform dimensionality reduction i.e. dig out a smaller set of features required for data classification
- Enforce model to learn by using available training data set
- Utilize trained model to categorize test/unknown data

This study defines a DLP model which uses various statistical approaches like TF-IDF (Term Frequency-Inverse Document Frequency) a renowned term count/weighing function, Vectorization, Gradient boosting document classification etc. [21] to classify the documents before accessing them. Proposed model also introduces an extremely efficient and more accurate approach IGBCA (Improvised Gradient Boosting Classification Algorithm); for document classification to prevent them from possible data leakage.

This study is organized as defined here: Section 2 provides a overview of work related to study; proposed data classification model is explained in Section 3; further Section 4 demonstrate the experimental work carried out in detail and provides an insight of research findings; Finally Section 5 depict conclusions made from study and also suggest future directions for further study.

## 2. Related Work

Numerous researchers have carried out a Wide study using different approaches like advanced fingerprinting, statistical analysis etc. and proposed various approaches for the sake of data security. Some of the research works related to proposed model are defined here. Shapira et al. [22] introduced advanced fingerprinting to overcome limitation of fingerprints by common data hashing. It uses k-skip-n-grams to generate advanced fingerprints as normal fingerprints can be easily compromised with little alterations in regular data. K-skip-n-grams approach can easily detect data even after alterations in data. But, this approach needs high storage and computations due to extreme indexing.

Hart et al. [23] discussed a Machine learning based method which categorizes organization documents as sensitive or non-sensitive. Support Vector Machines (SVMs) algorithm were utilized to categorize the organization private data, organization public data and non-organizational data. It used binary weighted unigrams to define the whole corpus. The major shortfall of this approach is, it just specify data as sensitive and non sensitive but unable to categorize data as secret, top secret or confidential class.

Salakhutdinov et al. [24] presented a TF-IDF based approach to detect the documents which have semantically similar content. An important limitation with which this approach suffers is that it can detect very generic documents only. Shu et al. [25] utilized sequence alignment techniques to detect data leaks due to transformations in data like insertion, deletion. It recognizes sensitive data patterns and uses a sampling algorithm to compare the resemblance between sensitive sequences. This system achieved high accuracy in detecting data leaks due to transformations. Alneyadi et al. [26] presented a DLP model based on statistical analysis to handle data which is fuzzy in nature. It can detect sensitive data and categorize it on basis of data semantics. Term Frequency-Inverse Document Frequency (TF-IDF) was used to categorize the data and SVD (Singular Value Decomposition) matrix to picture the results with high degree of precision and accuracy.

## 3. Proposed Model for Data Security

In this section complete end to end system details and architecture of proposed model as demonstrated in Figure 4 is explained and Figure 5 illustrate a simplified workflow of proposed document classification model to endow with data security; where text processing, BoW creation, Vectorization, TF-IDF and Gradient Boosting classification algorithm (GBCA) etc. approaches are utilized. This model proposes an Improvised Gradient Boosting classification algorithm (IGBCA) for model tuning. Most important contribution/gain of proposed model is the Improvised version of GBCA which provides better accuracy over existing GBCA approach.

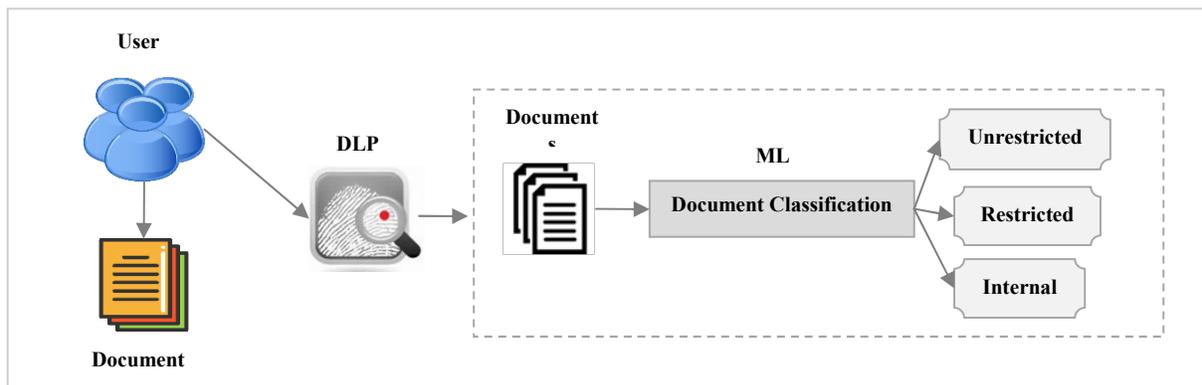

**Figure 4:** Proposed Model Architecture

### 3.1. Initial Pre Processing

The initial pre-processing involves creating a Reference data frame from the Label data. Subsequently based on this reference data, extract each document contents, which further leads to creation of another data frame to be used for training of document type classification model. The next move in initial preprocessing is to create another Data frame consisting of text from all documents in given repository. This data turn out to be the labeled data.

### 3.2. Text Pre Processing

Preprocessing is one of the most important steps while dealing with any kind of text models. Preprocessing is essential in topic modeling and is required to generate non-sensitive as well as sensitive words. Figure 6 highlights few mandatory preprocessing carried out on text: conversion to lowercase, punctuation removal, extracting stop words and lemmatization/stemming. Preprocessing is executed with the help of NLP/NLTK (Natural Language Processing) library. This results in creation of a list of sensitive words and check out sensitive words in extrovert network.

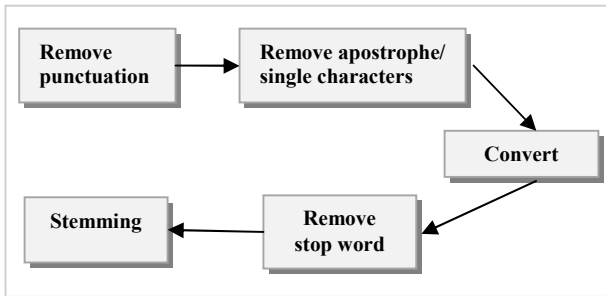

**Figure 6:** Text Preprocessing

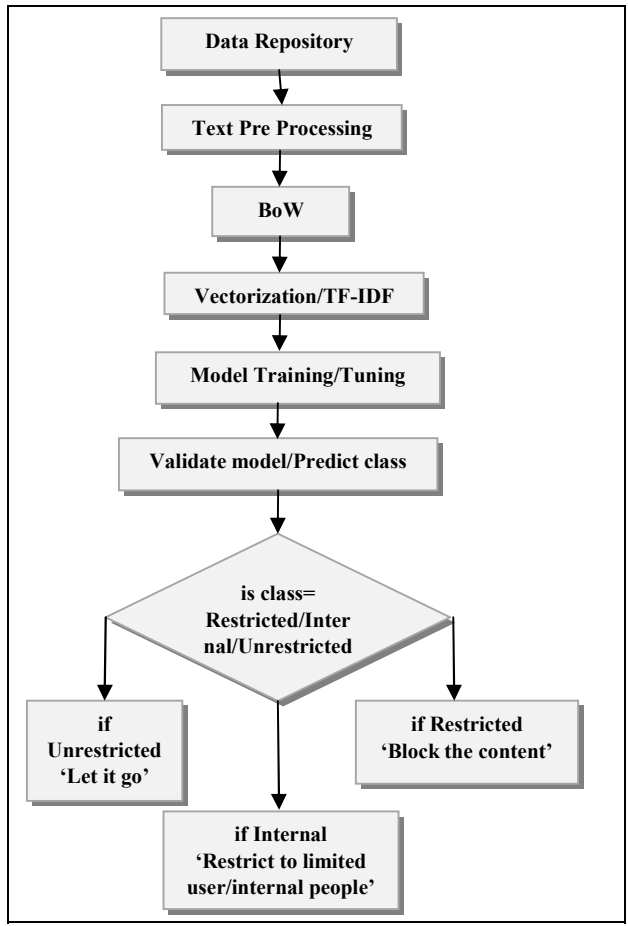

**Figure 5:** Layout of Document Classification Model

### 3.3. BoW Implementation

Post preprocessing all data becomes accessible in pure text format free from all kind of punctuations, repeating grammatical words, stop words etc. The data classification algorithm requires numerical vector to carry out the data classification. Therefore an essential need arises for conversion of the plain text into some sort of features. In order to drive features from existing textual data to reflect properties of textual data, feature extraction or feature encoding is requisite to perform. The BoW (Bag-of-Words) approach is most appropriate for the above discussed purpose; a text representation which describes all words existing within documents. BoW acts as a vocabulary of unique words occurring in documents. The entire features in BoW are known as list of Tokens/Lammas. BoW engrosses two things:

- A vocabulary of known words i.e. Lammas/Tokens
- A way to determine the presence of known words

Moreover, BoW don't posses any information regarding the order of occurrence of words in respective document; on a side it specifies whether a known word exist in document but at the other hand don't know where (position) it lies in document.

### 3.4. Train & Test data set

The following sub-section discusses the process to convert the available plain text data in numeric feature required for model training. As explained in above subsections, data classification algorithms need some vectorization (V) features to assign weight on each class. Converting the text into numeric feature is key part as this is an essential requirement to execute any machine learning model. Accordingly this sub-section focus on three major points to convert Lammas in numeric features.

- Find term frequency i.e. to obtain the count how many times a term/word occur in every document
- Grant less weight to frequently occurring tokens, process named as inverse document frequency
- Regularize the vectors i.e. perform normalization to obtain unit length vectors

### 3.4.1. Vectorization

The classification algorithms have need of numerical value to accomplish the assigned data classification task. As a result it becomes crucial to convert plain text into vector of numeric values. Actually a lot of methods exist to translate a corpus (of words) into required vector format. Data vectorization (V) is most suited approach, to drive numerical vector from existing textual data. Every unique word in the textual data is corresponded with a number. Thus all the raw textual data i.e. sequence of characters transformed in vectors i.e. sequences of numbers. It transforms all the raw textual data i.e. sequence of characters in the vectors i.e. sequences of numbers.

Dimensions in each vector are equal to the number of unique words in the whole corpus. Proposed model uses SciKit Learn's CountVectorizer to transform text documents collection into token count matrix. Table 1 below demonstrates a very simple 2D matrix in which one dimension represents the whole vocabulary as one row per word whereas the other dimension shows the documents as a column per text message. On same node it leads to creation of a sparse matrix; a matrix with maximum entries marked as zero. Sparse demonstration is difficult to illustrate due to computational complexity i.e. space as well as time complexity. Moreover it provides extraordinarily little information in such a bulky data representation space.

**Table 1:** An Example of a Vector Table

|                | Document 1 | Document 2 | … | Document N |
|---|---|---|---|---|
| **Word 1 Count** | 0 | 1 | … | 0 |
| **Word 2 Count** | 0 | 0 | … | 0 |
| **…** | 1 | 2 | … | 0 |
| **Word N Count** | 0 | 1 | … | 1 |

### 3.4.2. Word Scoring: TF-IDF

TF-IDF is a statistical method utilized to determine the importance of a word ($\omega$) for a document in the collection/corpus of data. The importance of word is proportionally related to the number of times a term/word ($\omega$) occur in particular document and is offset by the frequency of the term/word ($\omega$) in whole corpus. Importance is increasing in nature with respect to the frequency of term/word occurring. TF-IDF is product of ***tf*idf***.

At the first model compute the normalized Term Frequency (*tf*) as shown below in Eq. 1. Where, *tf* basically describe the frequency of a particular term/word ($\omega$) inside a document (D).

$$tf_i(\omega) = \left\{ \frac{\text{number of word } (\omega) \text{ appear in document}}{\text{total number of word in the document}} \right\} \quad (1)$$

The other term Inverse Document Frequency (*idf*) is an inverse measure to determine the in-formativeness about $\omega$. It is calculated as the log of *N* i.e. total number of documents appearing in the corpus divided to the number of documents containing that particular term/word.

$$idf_\omega = \log\left(N / Df_i\right) \quad (2)$$

Since documents may have different lengths, there is possibility that a particular term/word may occur more times in a long document than a short one. Due to this a need arise to align the data on same scale so that model biasness can be avoided. Therefore, the term frequency is divided with length of document to obtain normalized counts. Also some common least important terms like "is", "of", and "that" occur multiple times in document. Thus a need arise to scale up least frequently occurring terms and weigh down most frequent ones.

### 3.4.3. Feature Scaling

Now data is available in form of matrix of feature required to train the model. It is not recommended to train the model on all the features as fitting model on each and every feature may result in over fitting or less accuracy. Thus it becomes obligatory to select the important feature from the dataset upon which model is trained. Model uses Select K best method from Scikit learn package to identify the important features and train the model only on these important features selected. As a result feature scaling avoid the over fitting of the model and also helps to improve model accuracy.

### 3.5. Model Training

Gradient boosting uses a differential loss function for data classification. Results obtained are not good enough by using Gradient boosting algorithm. Therefore for model training improvised version of gradient boosting algorithm (IGBCA) is considered as the target outcome is based on the gradient of the error for each class and with respect to predicted value. Every model selects a direction where it can minimize the predicted error. As explained in above section it is difficult to achieve high accuracy of the model with limited source labeled data set. Therefore proposed scheme has applied scikit StratifiedKFold which first shuffle the data set and then splits the data into n splits parts. All these parts are used as a separate test set.

**Table 2:** Pseudo Code for Proposed Model

1. Perform K fold for cross validation
   paramGrid={
        (skf=StratifiedKFold(n_splits)),
        (SelectKBest(chi2))
        }
2. Create a pipeline of various functions as
   pipe= {
        (TfidfVectorizer()),
        (SelectKBest(chi2)),
        (model GBCAclassifier)
        }
3. Train/tune the model using random search
   model= { RandomizedSearchCV(pipe, paramGrid,
        cv=skf.split(x,y), scoring=accuracy)   }

### 3.6. Model Tuning

The proposed system aims to provide the high accuracy even with less dourer data set, hence improving the model accuracy is prime focus of the system. In process to improve the model accuracy the system become a part of ensemble learning. Current study uses Improvised Gradient Boosting Classification Algorithm (IGCBA) by introducing ensemble learning pipeline including Gradient Boosting Classifier algorithm (GBCA) explained above in Table 2, ParamGrid, kBest, StratifiedKFold and RandomizedSearchCV. All these libraries belong to scikit. Model provides incredibly fine accuracy by using this pipeline.

### 3.7. Model Validation

A way to validate the classification model is to compute some parameters like Sensitivity, Specificity, Recall, Precision and F-Measure, overall model accuracy and error rate.

**Sensitivity** ($\mathcal{S}$) tells how appropriate the model is to detect the positive class. Let's consider class 'Restricted' and check the sensitivity quantifies how correctly document class 'xyz' is predicted. Sensitivity ($\mathcal{S}$) is calculated as represented in Eq. 3.

$$\mathcal{S} = \frac{TP}{TP + FN} \qquad (3)$$

**Specificity** ($\mathfrak{S}$) explains how accurate negative class is mapped. It gives true negative rate. It can be calculated as represented by Eq. 4. It lies in range 0.0 to 1.0. 1.0 is considered as the best specificity ($\mathfrak{S}$) and worst is 0.0.

$$\mathfrak{S} = \frac{TN}{FP + TN}. \qquad (4)$$

**Precision** ($\rho$) measure how accurate model is assigning events means positive class to positive class. Precision is also known as error type2. It also plays a crucial role while assuring the correct class. Precision ($\rho$) score is computed by Eq. 5 as shown below:

$$\rho = \frac{TP}{TP + FP} \qquad (5)$$

**Recall** ($\mathfrak{R}$) is also explained similar to Sensitivity and is utilized for F-score calculation.

**F1-score** ($f$) is the harmonic mean (average) of the precision and recall. It is best suited for the training set having an uneven class distribution. Eq. 6 provides F1-Score.

$$f = \mathfrak{R}^* \rho / \mathfrak{R} + \rho \qquad (6)$$

Another parameter to inspect model performance is Overall model accuracy (Д) is computed as shown in Eq. 7

$$Д = \frac{TP + TN}{TP + TN + FN + FP} \qquad (7)$$

One more important parameter to validate the model is error rate calculation. **Error** ($\mathcal{E}$) is the difference between actual and predicted outcomes. Model has validated the accuracy by using errors as well as represented by Eq. 8.

$$\mathcal{E} = \frac{FP + FN}{TP + TN + FN + FP} \qquad (8)$$

## 4. Experimental Results & Discussions

Proposed study performance is computed on three different forms of document categories. Source data hold data class type restricted, internal and unrestricted. Model is validated on various parameters like Sensitivity, Specificity, Recall, Precision and F-Measure, overall model accuracy and error rate; as described in above section. Model train data set contain three distinct kinds of data; consequently **confusion matrix** is used to test the accuracy of the model. Figure 7 below demonstrates a confusion matrix; to validate the accuracy of the model based on sensitivity, Specificity, Recall, Precision and F-Measure. In case if model is 100% accurate then matrix would have been a diagonal matrix.

Computations provide sensitivity ($\mathcal{S}$) score 0.951 which reflects model has a very high accuracy of about 95%. Specificity ($\mathfrak{S}$) score value computed is 0.965; means proposed model evaluate 96% specificity score which is almost near to 1.0. Thus it can be considered that model is making good predictions as, specificity ($\mathfrak{S}$) lies in range 0.0 to 1.0 and 1.0 is believed as the best specificity ($\mathfrak{S}$). High **Precision** ($\rho$) is like assurance that correct class is determined. A very high Precision ($\rho$) 0.933 means model is able to predict 93% of 'Restricted' class as 'Restricted'.

**F1-score** ($f$) value is 0.943 which is approximately 1, this is possible when precision ($\rho$) and recall ($\mathfrak{R}$) both near to 1 i.e. system has predicted F1- score ($f$) which is comparable to precision and Recall. Overall model accuracy ($\text{Д}$) score falls near 0.960 based on this score the proposed system posses 96% accuracy which is very close to 1.0 i.e. a perfect model. A perfect model is extremely difficult to fit. Usually accuracy rate getting calculated in probability format as 1-error. Error ($\mathcal{E}$) rate falls near 0.0396 which appear to be very close to 0.0. Even if applied [1-error] formula still accuracy is 96%.

If model predict class 'Restricted' then that document/content should not be passed out of premises or can be blocked by system. If model classify document as 'internal', it means this document/content can flow within the organization premises and no access out of premises. And if model specifies document as 'unrestricted' then it is up to organization whether to block this or let it go out of organization. Based on all the above computations and their outcomes the proposed system can be considered good enough and an effective solution towards the DLP for data security. Therefore model is highly recommendable for the data security purpose. It is highly suitable for data prevention from unauthorized entities.

|  |  | POSITIVE | NEGATIVE | NEGATIVE |  |
|---|---|---|---|---|---|
| **Model Predictions** | POSITIVE Actual class (Restricted) | 98 | 2 | 3 | FN |
|  | NEGATIVE Actual class (Internal) | 5 | 95 | 0 |  |
|  | NEGATIVE Actual class (Unrestricted) | 2 | 1 | 97 |  |
|  |  | FP | TN | | |

**Figure 7:** Confusion Matrix

## 5. Conclusion and Future Scope

This paper proposes a classic text classification technique using machine learning based statistical analysis approach which classifies documents to prevent from data leak as a data leakage prevention solution. It uses well known BoW technique for feature extraction and highly renowned retrieval function TF-IDF for term weight/count. Model uses improvised gradient boosting classification algorithm (IGBCA) for model training/tuning. Proposed model achieved high score/value of various parameters like sensitivity, specificity, F1-score and precision which proves efficiency of model. Conclusion drawn on the basis of study is that proposed model is performing good enough to be considered as an efficient and effective DLP solution. Speed of classification and various other system overheads are most important vulnerable factors need to be tackled in future.